\documentclass{article}
\usepackage{spconf,amsmath,epsfig}
\usepackage{graphicx}
\usepackage{url}
\usepackage{inputenc}
\usepackage{textcomp}
\usepackage{subcaption}
\let\OLDthebibliography\thebibliography
\renewcommand\thebibliography[1]{
  \OLDthebibliography{#1}
  \setlength{\parskip}{0pt}
  \setlength{\itemsep}{0pt plus 0.3ex}
}

\begin{document}\sloppy

\def\x{{\mathbf x}}
\def\L{{\cal L}}

\title{A Single RGB Camera Based Gait Analysis with a Mobile Tele-Robot for Healthcare}
%
\name{Ziyang Wang\textsuperscript{1}\thanks{Corresponding Author ziyang.wang@cs.ox.ac.uk}, Fani Deligianni\textsuperscript{2}, Qi Liu\textsuperscript{1}, Irina Voiculescu\textsuperscript{1}, Guang-Zhong Yang\textsuperscript{3}}
\address{\textsuperscript{1}Department of Computer Science, University of Oxford\\
\textsuperscript{2}School of Computing Science, University of Glasgow\\
\textsuperscript{3}Institute of Medical Robotics, Shanghai Jiao Tong University}

\maketitle

\begin{abstract}
With the increasing awareness of high-quality life, there is a growing need for health monitoring devices running robust algorithms in home environment. Health monitoring technologies enable real-time analysis of users' health status, offering long-term healthcare support and reducing hospitalization time. The propose of this work is twofold, the software focuses on the analysis of gait, which is widely adopted for joint correction and assessing any lower limb, or spinal problem. On the hardware side, a novel marker-less gait analysis device using a low-cost RGB camera mounted on a mobile tele-robot is designed. As gait analysis with a single camera is much more challenging compared to previous works utilizing multi-cameras, a RGB-D camera or wearable sensors, we propose using vision-based human pose estimation approaches. More specifically, based on the output of state-of-the-art human pose estimation models, we devise measurements for four bespoke gait parameters: inversion/eversion, dorsiflexion/plantarflexion, ankle and foot progression angles. We thereby classify walking patterns into normal, supination, pronation and limp. We also illustrate how to run the proposed machine learning models in low-resource environments such as a single entry-level CPU. Experiments show that our single RGB camera method achieves competitive performance compared to multi-camera motion capture systems, at smaller hardware costs.
\end{abstract}

\section{Introduction}

\begin{figure*}[h]
    \centering
        \includegraphics[width=6in]{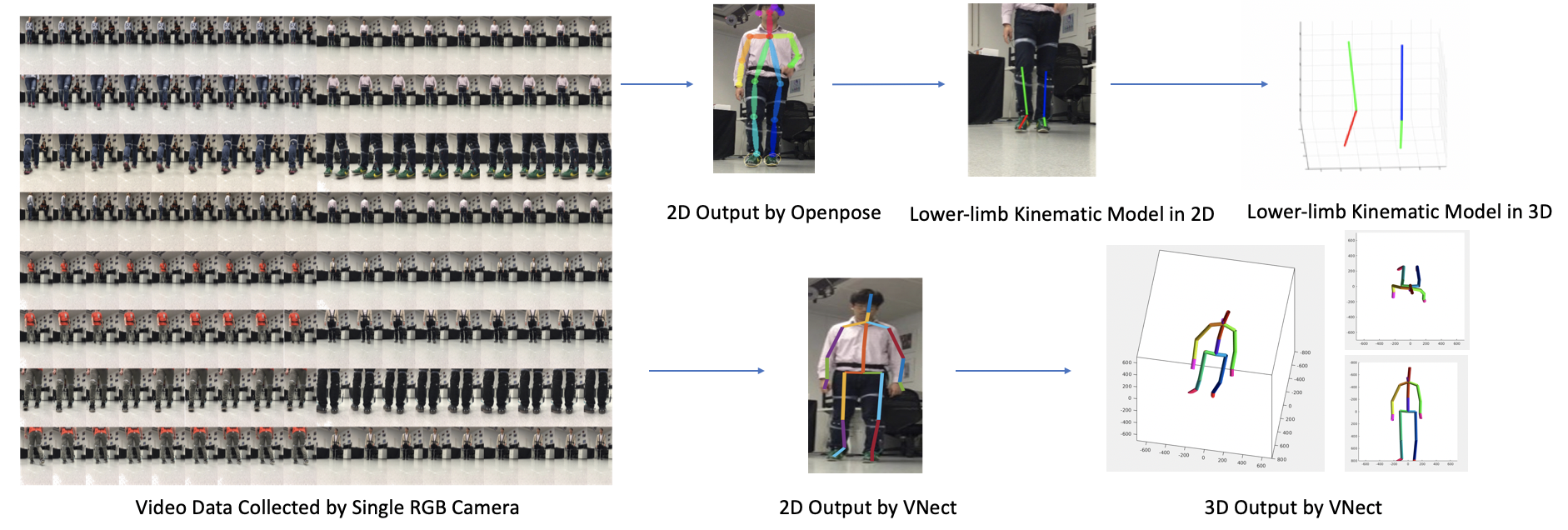}
        \caption{The Example Output for Gait Analysis Based on VNect And Openpose}
        \label{fig:example}
\end{figure*}

New technologies which monitor users in a non-clinical environment have been growing in popularity. They keep users engaged and encourage proactive healthcare, providing long-term medical support. Among health monitoring tasks, gait disorders are very common in elderly, patients after surgery or sports injury, which influences the quality of life. This work focuses on gait analysis, which measures user walking patterns for joint correction and assessment of lower limb or spinal problem. A noteworthy application of gait analysis is home monitoring of the gait of patients with neurological disorders, which allows for early intervention and reduces medical expenses. Previous works apply depth cameras \cite{rgbd} or optical markers \cite{optical}, both of which imply a fairly high cost and limit real-life applicability. Motivated by the recent advances in computer vision, especially human pose estimation and its wide applications in biomedical research, gaming industry and surveillance \cite{humanposeestimationreview}, our design involves a novel marker-less gait analysis device with a low-cost RGB camera mounted on a mobile tele-robot. A single-camera-based gait analysis has two main challenges: (A) Depth ambiguities, motion blurs, unconstrained motions and self-occlusion are harmful for gait prediction \cite{Fani1}. (B) A single image can usually be mapped to more than two possible human poses \cite{humanposeestimationreview}. In order to circumvent these challenges, we use two state-of-the-art 2D and 3D human pose estimation algorithms, Openpose \cite{Openpose} and VNect \cite{Vnect}. We devise measurements for four bespoke gait angular parameters of the lower limb: inversion/eversion, dorsiflexion/plantarflexion, ankle and foot progression angles. These are defined, extracted and measured. We thereby classify walking patterns into normal, supination, pronation and limp based on these gait features. 

Multi-camera systems are not easy to deploy in real life environments. By contrast, our single camera system has been developed as an autonomous mobile tele-robot to be used in a home-like environment. Machine learning models have been implemented in such a way as to run on a single entry-level CPU, such as iPad. An iOS APP with a dedicated user interface, where the skeleton of users in front of the camera is shown in real time, is also developed. As well as processing human gait data, the APP also serves as a tele-robot controller and the robot functions well in an uncluttered environment.

We evaluated the proposed method using manually labelled 2D and 3D ground truth data. Our method achieves competitive performance compared to a commercial multi-camera motion-capturing system, while still requiring lower hardware costs. The code for the iOS APP has been made public through Github.

\section{Methods \label{sec:method}}

\subsection{From Human Pose Estimation to Gait Analysis}
Both VNect \cite{Vnect} and Openpose \cite{Openpose} are real-time methods to capture skeletal points of human with 2D images. 

Openpose applies a bottom-up method of person detection and pose estimation that detects and tracks 17 skeletal feature points based on the Part Affinity Field (PAF). The model can track multiple persons in real time. Figure \ref{fig:example} shows an example of Openpose used for gait analysis. To detect the key points of toes and get the orientation of feet, Gaussian mixture model-based image segmentation is applied \cite{Xiaoa,Grabcut}. The 2D key points of the joints are mapped to 3D space using active shape modelling together with the dictionary learning \cite{Xiaoa}.

VNect estimates human pose through tracking 21 global skeletal points in 3D, including the foot area. To speed up training and inference, it uses a 50-layer convolutional network. Kinematic skeleton constraints are introduced to improve stability and temporal consistency. Taking the temporal history of the sequence as input, the 2D locations of each joint from the 2D heatmap are transferred to a few links. Then, the length of each link, which is the most important reference in kinematic skeleton constraints, are calculated. Finally, the predictions from the 2D heatmap are mapped to 3D global poses. Only a single person can be tracked with VNect in real-time. Figure \ref{fig:example} shows an example output of VNect, where the 2D skeletal image is drawn based on the heatmap, and the skeletal image in 3D global space.

Several gait analysis parameters are specially extracted and processed. As normal and pathological gaits exhibit varied patterns, six key joints including left and right knees, ankles, toes and their corresponding angles are studied. Inversion/eversion angle, dorsiflexion/plantarflexion angle, and ankle angle are used to evaluate the proposed method's performance. The foot progression angle is studied as a gait angular feature to recognise gait patterns. For example in the $i_{th}$ frame, the left knee, ankle and toe as three joints are represented  $\alpha_{Li}=(xk_{Li},yk_{Li},zk_{Li})$, $\beta_{Li}=(xa_{Li},ya_{Li},za_{Li})$, $\gamma_{Li}=(xt_{Li},yt_{Li},zt_{Li})$. The left shank and foot as two links are represented $\Vec{S}_{Li}=\beta_{Li}-\alpha_{Li},\Vec{F}_{Li}=\beta_{Li}-\gamma_{Li},$
Inversion/eversion angle is the angle between the foot and horizontal plane. Dorsiflexion/plantarflexion angle is the angle between foot and vertical plane.
Ankle angle is the angle between foot and shank. The angular e.g. $A_{Li}$ is illustrated in equation \ref{1}.
\begin{equation}
    A_{Li}=\arccos{\frac{\Vec{C}_{Li}\cdot\Vec{F}_{Li}}{||\Vec{C}_{Li}||\times||\Vec{F}_{Li}||}}\label{1}
\end{equation}
Foot progression angle is the angle between the line e.g. $\Vec{F}_{Li}$ from the calcaneus to the second metatarsal and the line P as $\frac{x-x_{0}}{m}=\frac{y-y_{0}}{n}=\frac{z}{1}$ of progression averaged from heel strike to toe off during the stance phase of walking for each step. P is illustrated in equation \ref{2}, and calculated by equation \ref{3}.
\begin{equation}
    \begin{bmatrix} m & x_{0} \\ n & y_{0} \end{bmatrix}\begin{bmatrix} z  \\ 1 \end{bmatrix} = \begin{bmatrix} x \\ y \end{bmatrix}\label{2}
\end{equation}

\begin{equation} 
    \begin{bmatrix} m & x_{0} \\ n & y_{0} \end{bmatrix}=\begin{bmatrix} \sum xa_{i} za_{i} & \sum xa_{i} \\ \sum ya_{i}  za_{i}  & \sum ya_{i}  \end{bmatrix} \begin{bmatrix} \sum za_{i}^2  & \sum za_{i} \\ \sum za_{i} & n \end{bmatrix}^{-1}\label{3}
\end{equation}

\begin{figure}[h]
    \centering
    \begin{subfigure}[b]{0.8\linewidth}
        \includegraphics[width=\linewidth]{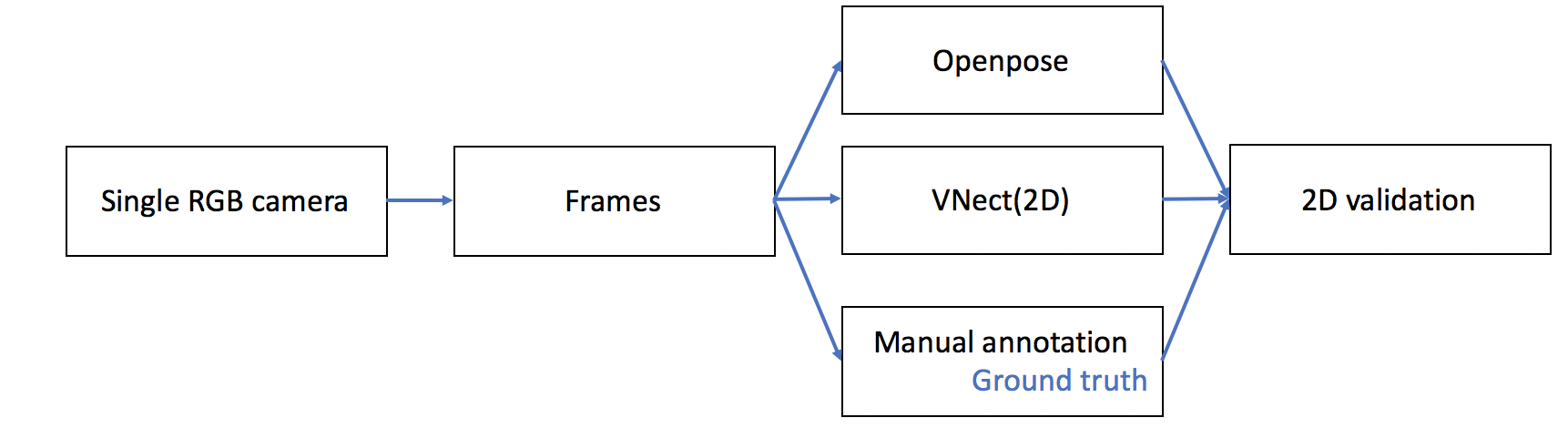}
        \caption{A work flow of 2D validation}
        \label{fig:framework1}
    \end{subfigure}
     
    \begin{subfigure}[b]{0.8\linewidth}
        \includegraphics[width=\linewidth]{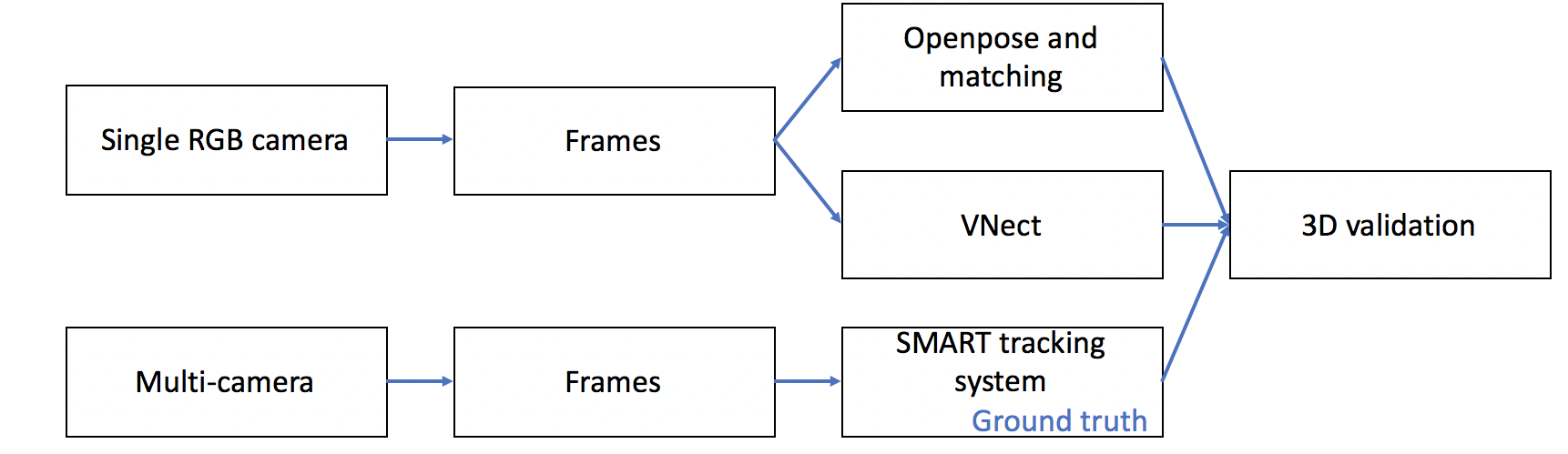}
        \caption{A work flow of 3D validation}
        \label{fig:framework2}
    \end{subfigure}
    \caption{Framework of validation}\label{fig:Frameworkvalidation}
\end{figure}

The evaluation experiment was carried out in a lab of 6*6 meters, equipped with a multi-camera acquisition system (SMART-DX) where markers are attached on human bodies. The multi-camera system was used in order to generate ground truth data. Simultaneously, one of the corners of the lab hosted a low-cost single RGB camera system, which collected raw walking data for validation.
Four volunteers were asked first to walk naturally and then to imitate three abnormal walking styles: limping, pronating and supinating. In order to synchronize the data between the single camera and the multi-camera system, the volunteers were asked to jump before and after walking. Both 2D and 3D ground truth data were collected by manually annotating the stream from the multi-camera system. A validation pipeline for each approach is shown in Figure \ref{fig:Frameworkvalidation}, the differences between the estimated results and the ground truth results are calculated as angular errors of the evaluation metric. As a result of the classification network, four separate walking patterns are accurately classified based on different distributions of gait features.

\subsection{System Design and Deployment}
\begin{figure}[h]
    \centering
        \includegraphics[width=0.7\linewidth]{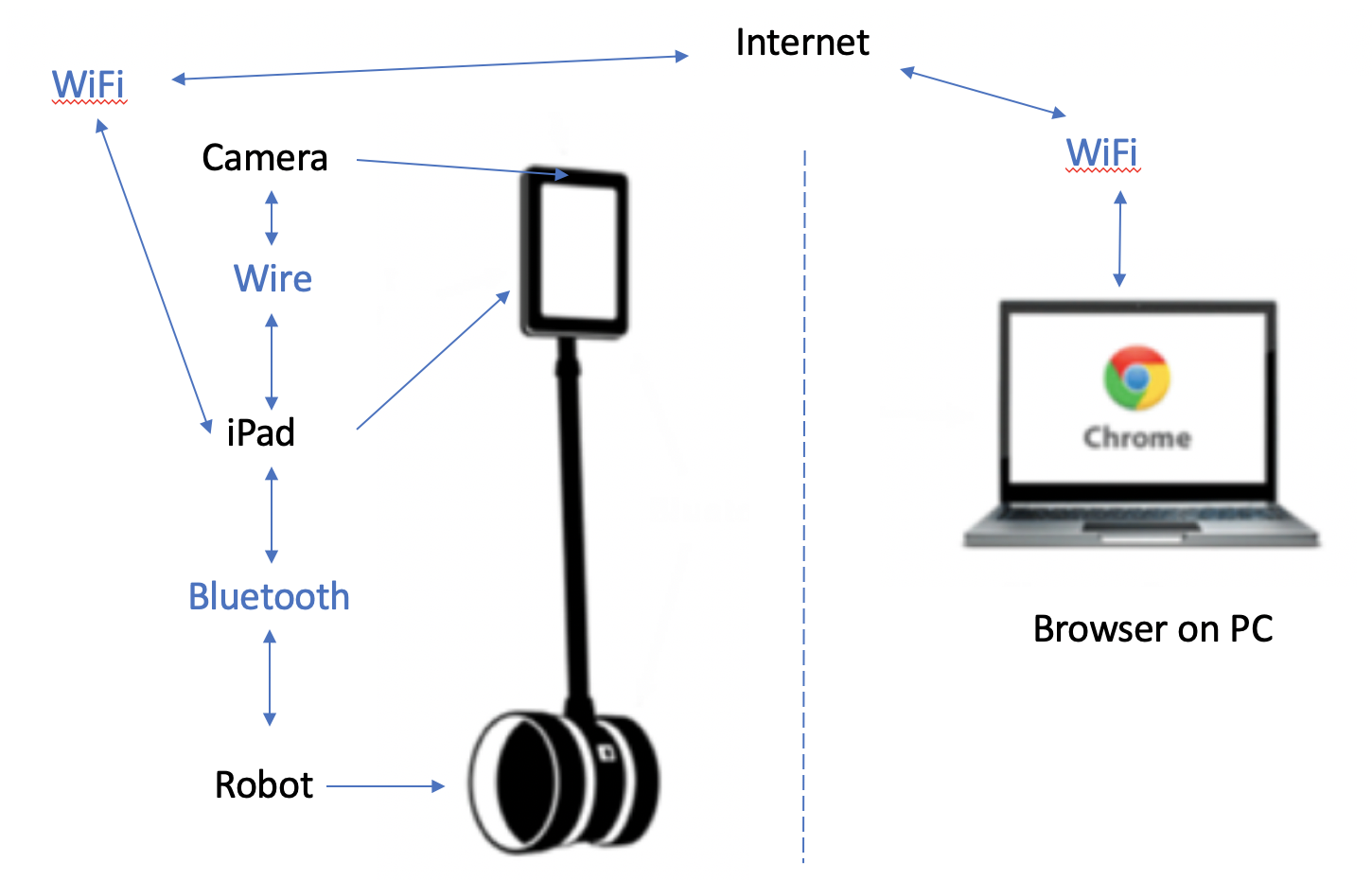}
        \caption{The framework of how Double Robot works}
        \label{fig:Demo4}
\end{figure}

Having tested and validated the software component of our system, we were keen also to realize its real-life application. Tele-robots are semi-autonomous robots which can be controlled remotely using wireless networks such as WiFi or Bluetooth. Figure \ref{fig:Demo4} illustrates our setup using a Double Robot, an off-the-shelf telepresence robot for iPad tablets, with a single RGB camera. It supports remote control, real-time video chat, auto-answering and self-balancing.



\begin{figure}[h]
    \centering
        \includegraphics[width=\linewidth]{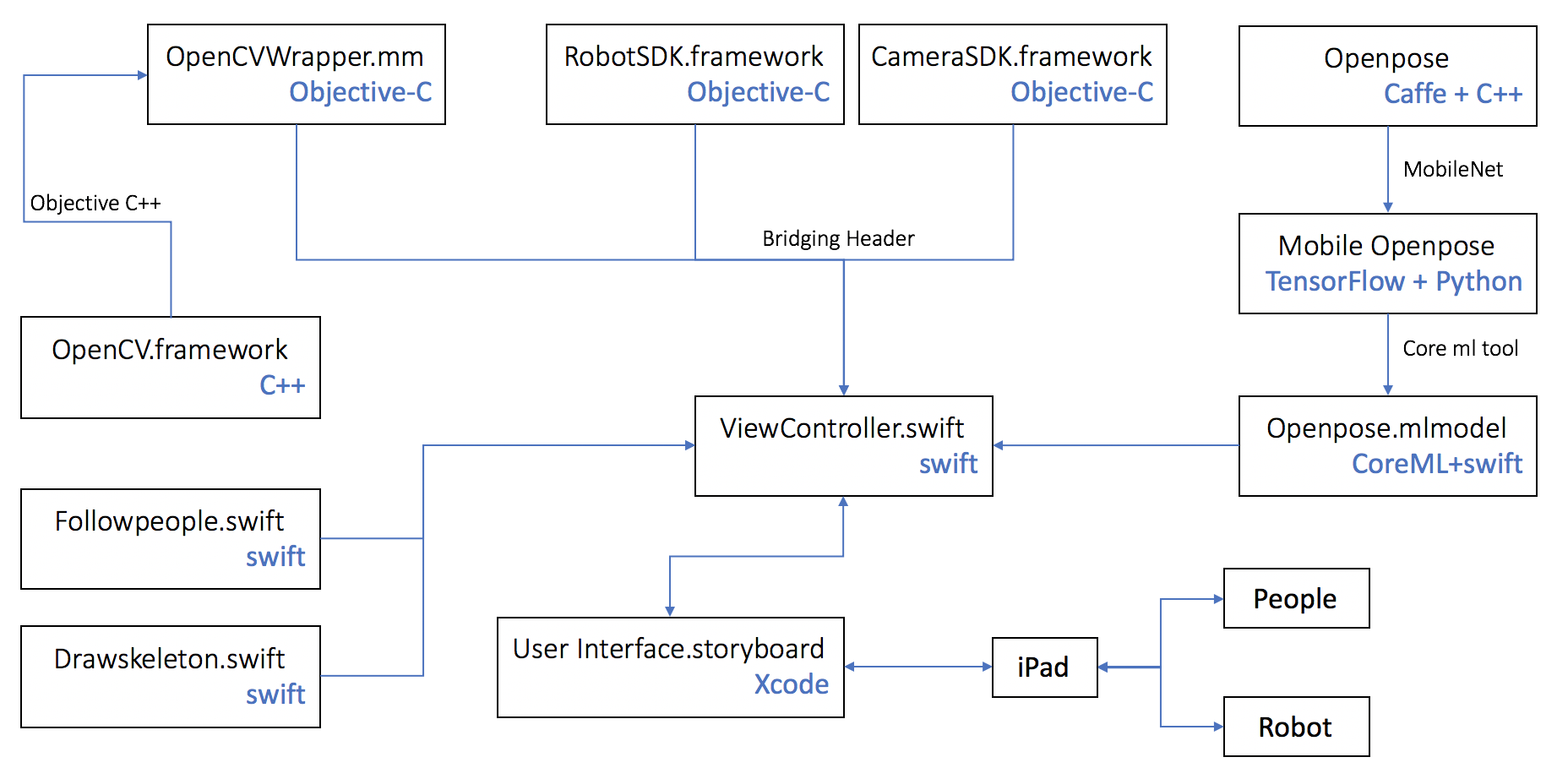}
        \caption{The framework of software development for iOS APP}
        \label{fig:Framework3}
\end{figure}

The software framework is shown in Figure \ref{fig:Framework3}. Since the Openpose model is computationally expensive, we modified it using MobileNet \cite{MobileNet} to accelerate the model processing. The modified model in the Caffe format is then converted to the CoreML format, which is applicable in an iOS platform. The frame rate achieves 3.5 in the 5th iPad whereas the original model running on Macbook Pro is only 0.6. Since the iPad was connected to the Internet through WIFI and controlled robot with Bluetooth, a remote control system was developed using the UDP protocol.


\section{Results}
\begin{figure}[h]
    \centering
    \begin{subfigure}{0.41\linewidth}
        \includegraphics[width=\linewidth]{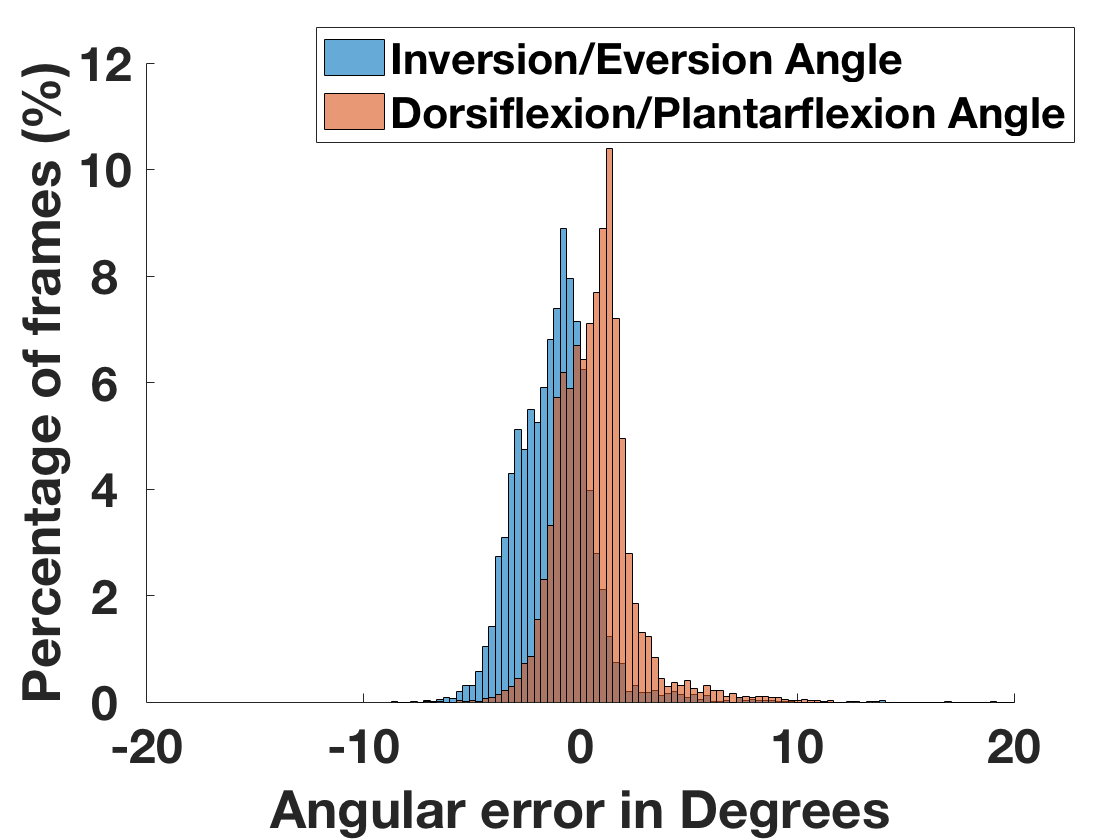}
        \caption{}
        \label{fig:Openpose2d1}
    \end{subfigure}
    ~ 
    \begin{subfigure}{0.41\linewidth}
        \includegraphics[width=\linewidth]{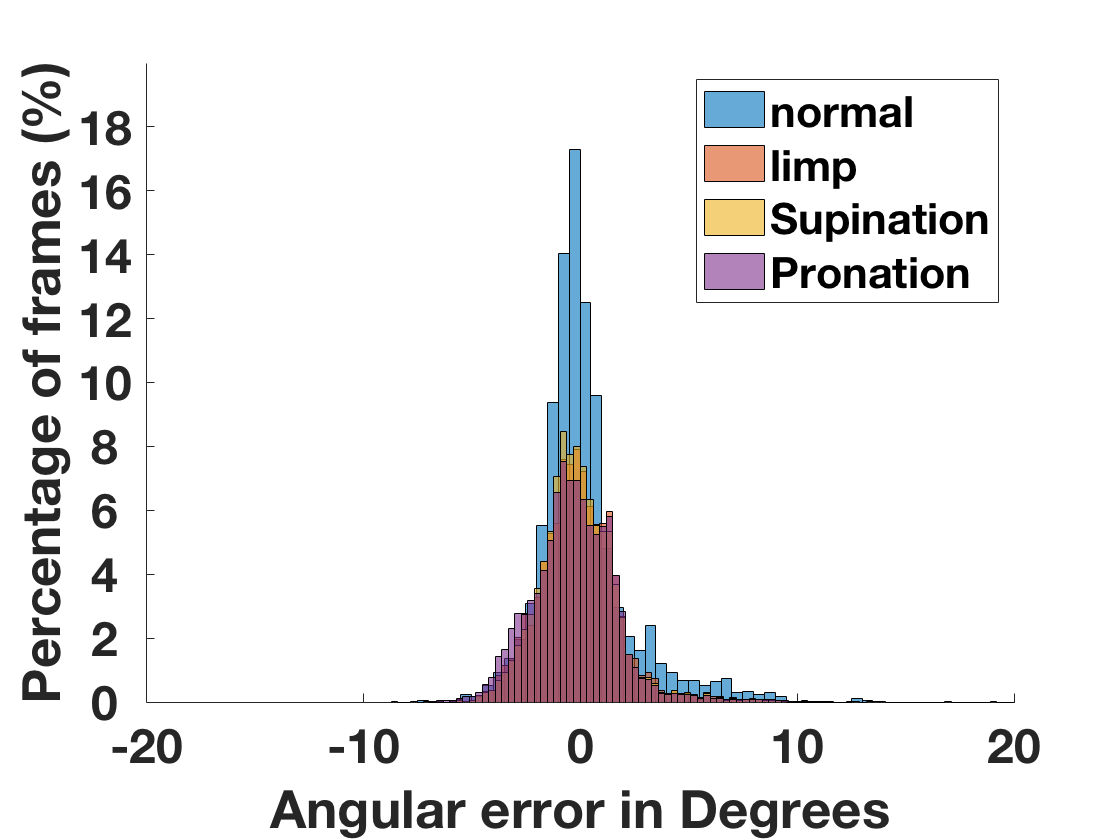}
        \caption{}
        \label{fig:Openpose2d2}
    \end{subfigure}

    \begin{subfigure}{0.41\linewidth}
        \includegraphics[width=\linewidth]{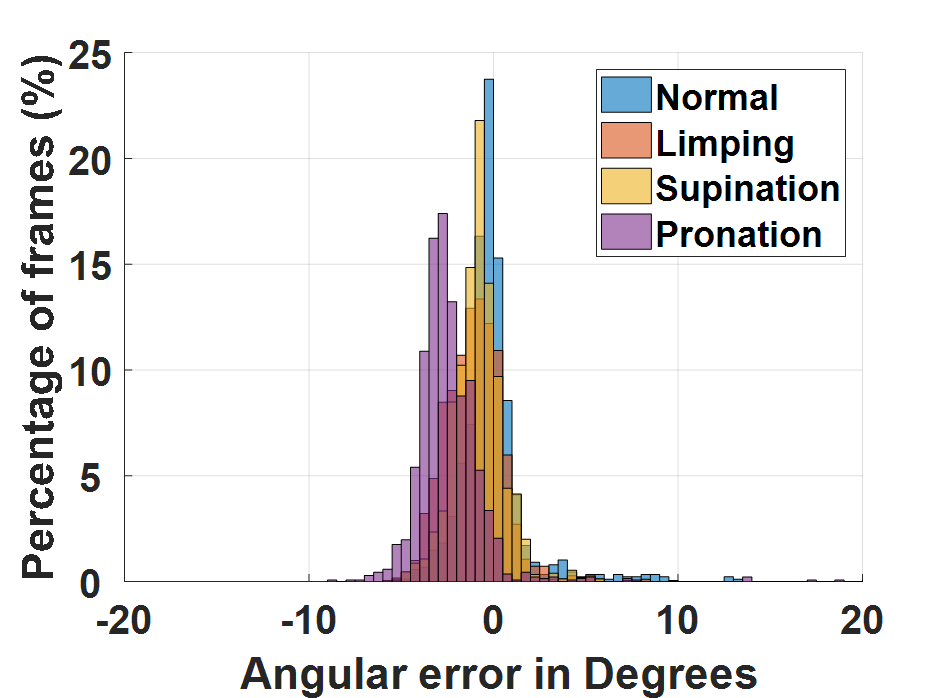}
        \caption{}
        \label{fig:Openpose2d3}
    \end{subfigure}
    ~
        \begin{subfigure}{0.41\linewidth}
        \includegraphics[width=\linewidth]{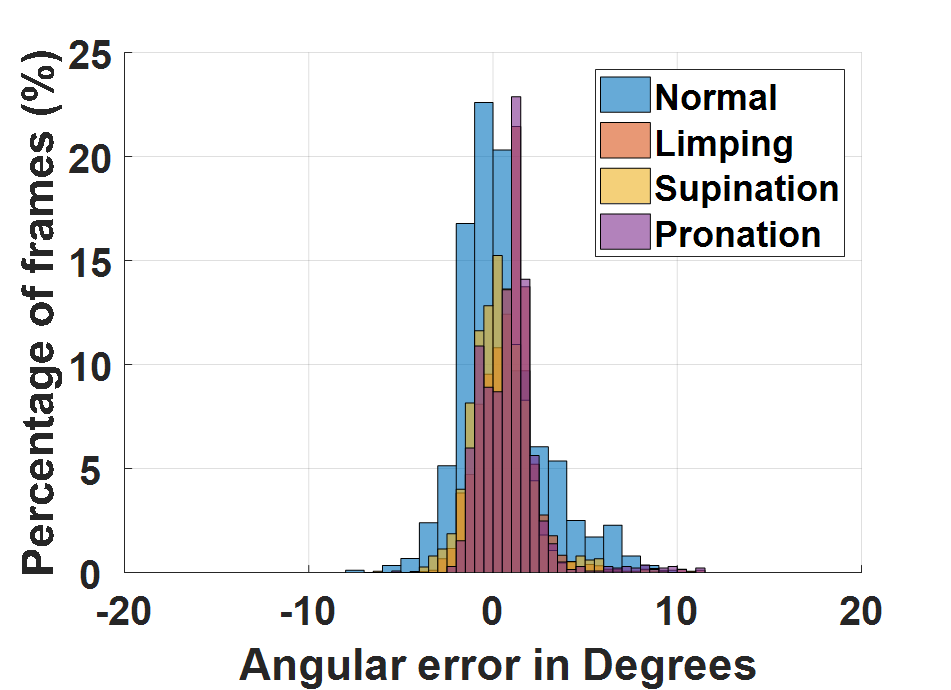}
        \caption{}
        \label{fig:Openpose2d4}
    \end{subfigure}
    
        \begin{subfigure}{0.41\linewidth}
        \includegraphics[width=\linewidth]{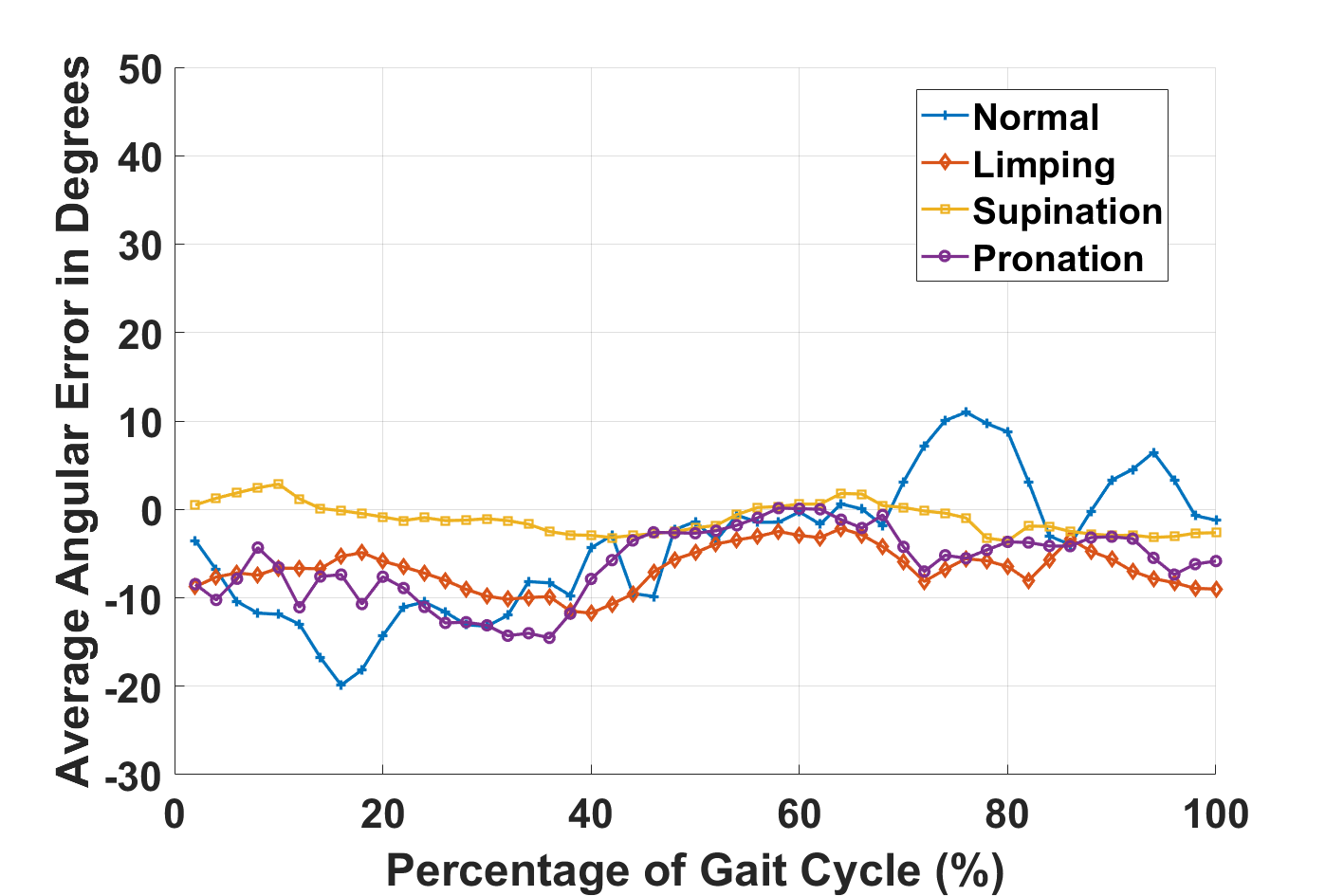}
        \caption{}
        \label{fig:Openpose3d1}
    \end{subfigure}
    ~
        \begin{subfigure}{0.41\linewidth}
        \includegraphics[width=\linewidth]{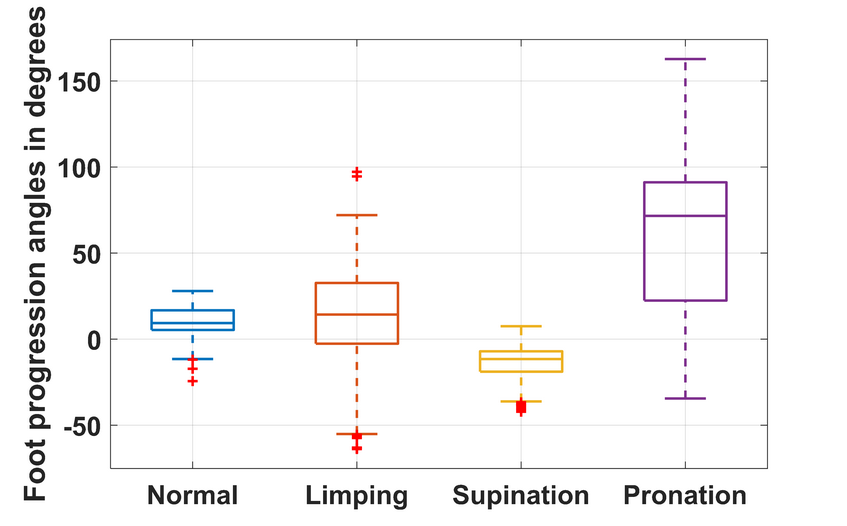}
        \caption{}
        \label{fig:Openpose3d2}
    \end{subfigure}
    \caption{Evaluation and Angular Features by Openpose}\label{fig:Openposeresults}
\end{figure}

\begin{figure}
    \centering
    \begin{subfigure}[h]{0.41\linewidth}
        \includegraphics[width=\linewidth]{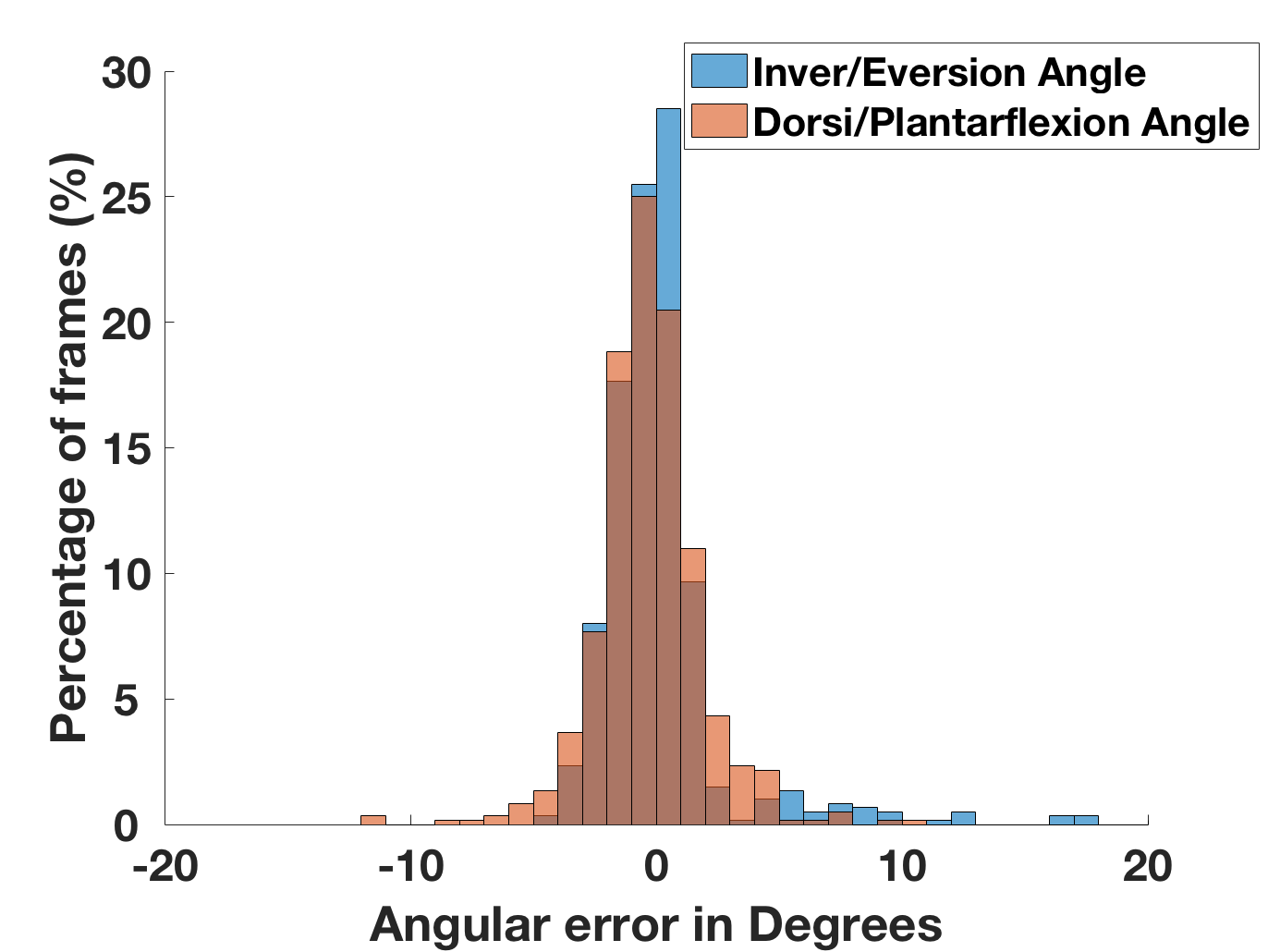}
        \caption{}
        \label{fig:Vnect2d1}
    \end{subfigure}
    ~ 
    \begin{subfigure}[h]{0.41\linewidth}
        \includegraphics[width=\linewidth]{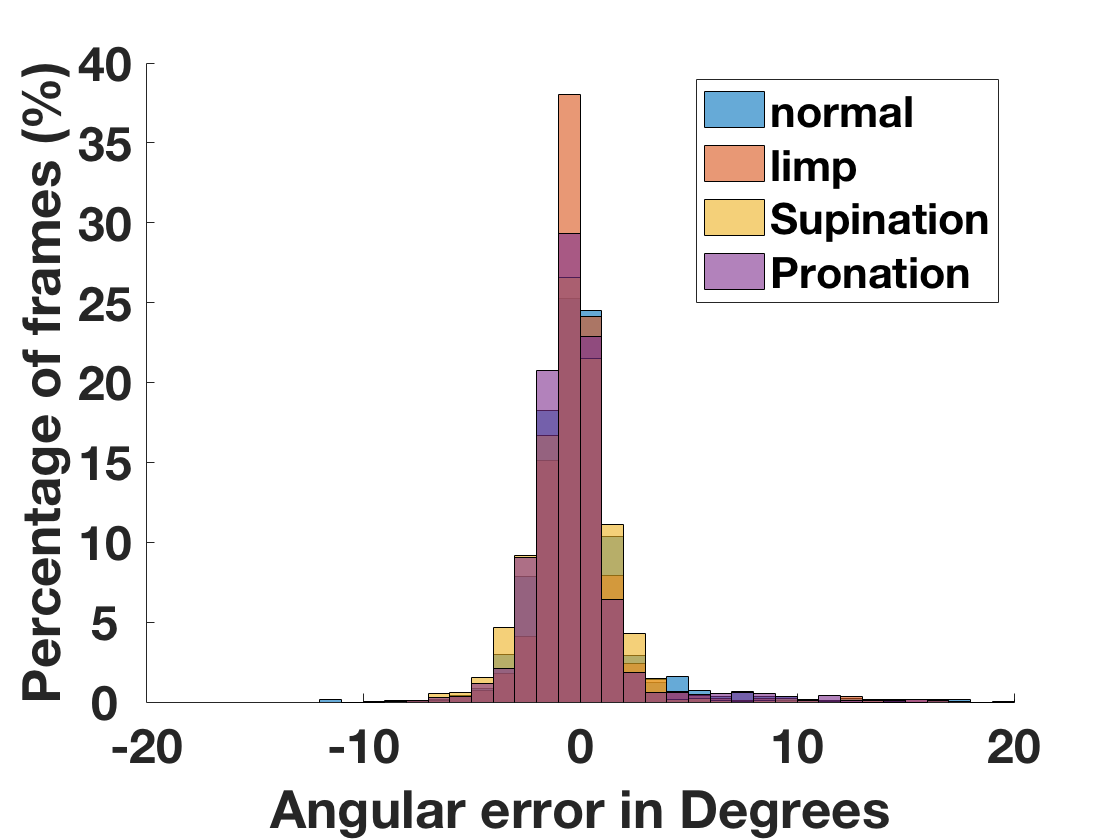}
        \caption{}
        \label{fig:Vnect2d2}
    \end{subfigure}

    \begin{subfigure}[h]{0.41\linewidth}
        \includegraphics[width=\linewidth]{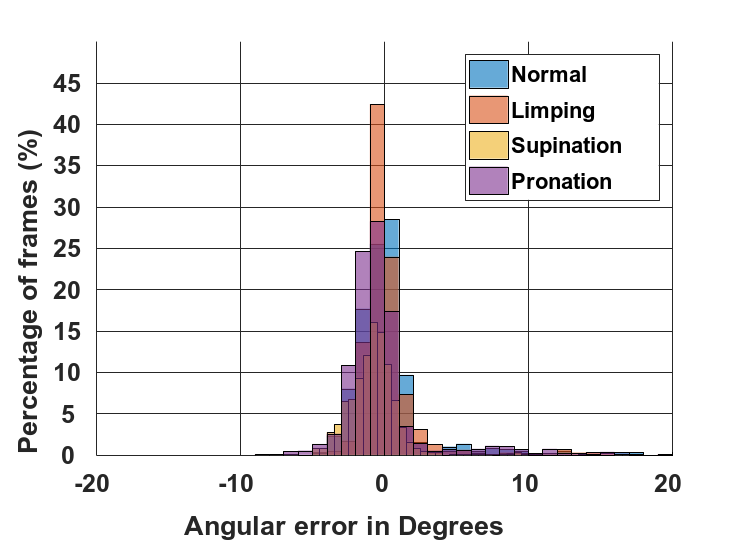}
        \caption{}
        \label{fig:Vnect2d3}
    \end{subfigure}
    ~
        \begin{subfigure}[h]{0.41\linewidth}
        \includegraphics[width=\linewidth]{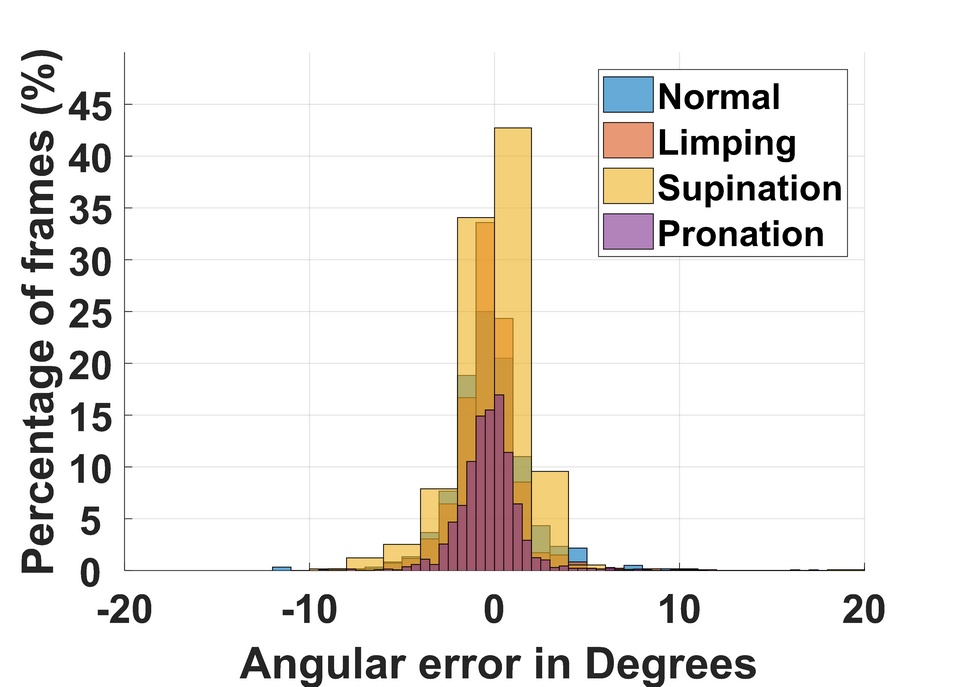}
        \caption{}
        \label{fig:Vnect2d4}
    \end{subfigure}
    
        \begin{subfigure}[h]{0.41\linewidth}
        \includegraphics[width=\linewidth]{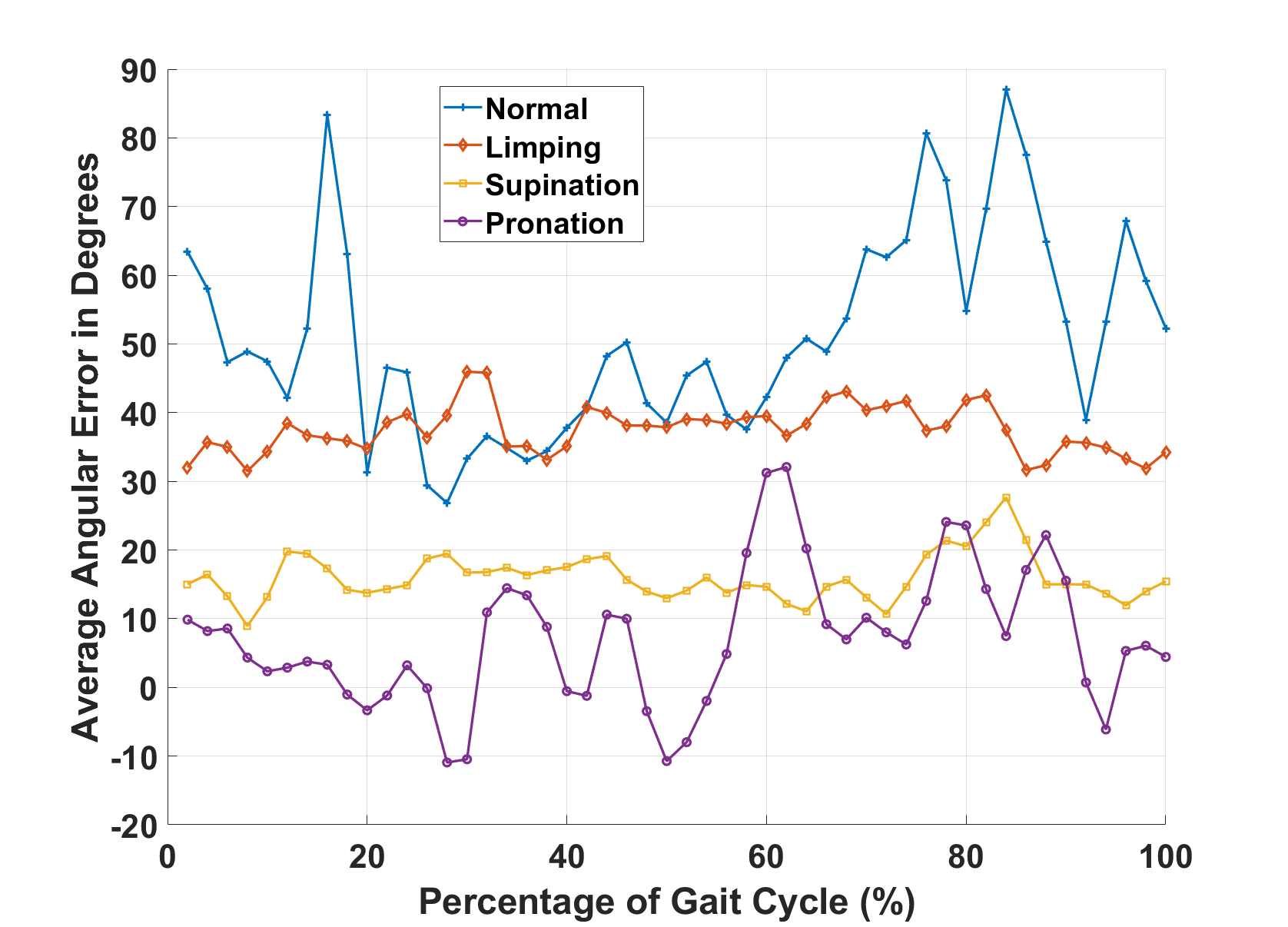}
        \caption{}
        \label{fig:Vnect3d1}
    \end{subfigure}
    ~
        \begin{subfigure}[h]{0.41\linewidth}
        \includegraphics[width=\linewidth]{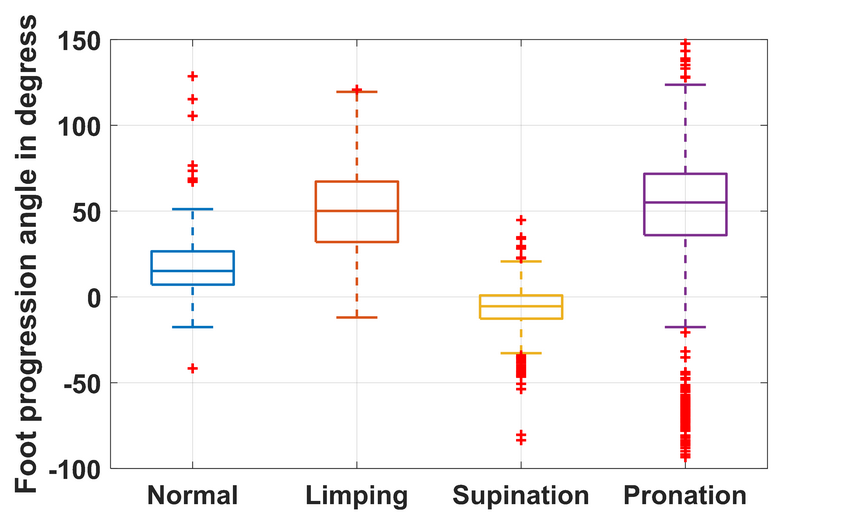}
        \caption{}
        \label{fig:Vnect3d2}
    \end{subfigure}
    \caption{Evaluation and Angular Features by VNect}\label{fig:Vnectresults}
\end{figure}

As the machine learning model has shown the best performance, the evaluation work in this paper focus on whether it could be utilised for gait analysis and comparison between a low-cost single RGB camera and commercial multi-camera with markers. Each of the dorsiflexion/plantarflexion and inversion/eversion angles measured by our system were compared against the values obtained from manual annotations and multi-camera. Any angular difference from the ground truth was considered to be an error. A total of 21,210 frames were studied. The comparisons are shown in Figure \ref{fig:Openposeresults} and \ref{fig:Vnectresults}. 

For 2D validation, we use histograms where the X-axis indicates angular error in degrees and the Y-axis is the percentage of frames where such error occurs. For 3D validation, we use line charts where the X-axis shows the percentage of the gait cycle and the Y-axis is average angular error in degree. The estimation results are shown in box charts which illustrate clearly the distribution of angular features. Figure \ref{fig:Openpose2d1} and \ref{fig:Vnect2d1} show the total 2D angular errors of inversion/eversion and dorsiflexion/plantarflexion. Figure \ref{fig:Openpose2d2} and \ref{fig:Vnect2d2} show the total 2D angular errors for each condition. Figure \ref{fig:Openpose2d3}, \ref{fig:Vnect2d3}  , \ref{fig:Openpose2d4} and \ref{fig:Vnect2d4} show the 2D angular errors for inversion/eversion angle and dorsiflexion/plantarflexion angle for four walking patterns, respectively. Figure \ref{fig:Openpose3d1} and \ref{fig:Vnect3d1} present the average 3D foot-leg angular errors in a gait cycle. Figure \ref{fig:Openpose3d2} and \ref{fig:Vnect3d2} present the foot progression angle distribution across different walking conditions.

For 2D angular features, VNect is slightly better than Openpose. The reason is that the toe point is estimated in different ways. VNect model uses toe points during training, but we have to apply ellipse fitting and image segmentation methods to estimate toe points for OpenPose, which results in error propagation. The 3D accuracy of VNect is substantially lower than modified Openpose-based approach. We conjecture that the experimental conditions (camera angle/walking route) do not meet optimal requirements for the 3D kinematic model of VNect. Furthermore, the model does not perform well when the upper body is occluded, which occurs when subjects come very close to the camera. Openpose perform on par with VNect on the foot progression angular features.


\section{Discussion}
We investigated vision-based gait analysis with a single RGB camera. We evaluated the accuracy of two state-of-the-art human pose estimation methods and validated them compared to a 3D multi-camera system. We also summarize the gait angular features of four different types of gait. Vision-based pose estimation requires low-cost camera but high computing power. Modification for machine learning models and porting work to an embedded system is completed. A demo mobile robot was further developed for real-life applications and be operated smoothly, widely and stably. The future work should be concentrated on the accuracy of human lower limb estimation and improving robot performance.


\bibliographystyle{IEEEbib}
\bibliography{icme2020template}

\begin{thebibliography}{1}

\bibitem{rgbd}
Yao Guo, Fani Deligianni, and etc,
\newblock ``3-d canonical pose estimation and abnormal gait recognition with a
  single rgb-d camera,''
\newblock {\em IEEE Robotics and Automation Letters}, vol. 4, no. 4, pp.
  3617--3624, 2019.

\bibitem{optical}
Paavo Vartiainen, Timo Bragge, Jari~P. Arokoski, and Pasi~A. Karjalainen,
\newblock ``Nonlinear state-space modeling of human motion using 2-d marker
  observations,''
\newblock {\em {IEEE} Trans. Biomed. Engineering}, vol. 61, no. 7, pp.
  2167--2178, 2014.

\bibitem{humanposeestimationreview}
Nikolaos Sarafianos, Bogdan Boteanu, and etc,
\newblock ``3d human pose estimation: {A} review of the literature and analysis
  of covariates,''
\newblock {\em Computer Vision and Image Understanding}, vol. 152, pp. 1--20,
  2016.

\bibitem{Fani1}
Fani Deligianni, Charence Wong, Benny Lo, and Guang{-}Zhong Yang,
\newblock ``A fusion framework to estimate plantar ground force distributions
  and ankle dynamics,''
\newblock {\em Information Fusion}, vol. 41, pp. 255--263, 2018.

\bibitem{Openpose}
Zhe Cao, Tomas Simon, and etc,
\newblock ``Realtime multi-person 2d pose estimation using part affinity
  fields,''
\newblock pp. 1302--1310, 2017.

\bibitem{Vnect}
Dushyant Mehta, Srinath Sridhar, and etc,
\newblock ``Vnect: real-time 3d human pose estimation with a single {RGB}
  camera,''
\newblock {\em {ACM} Trans. Graph.}, vol. 36, no. 4, pp. 44:1--44:14, 2017.

\bibitem{Xiaoa}
X.~Gu, Fani Deligianni, and etc,
\newblock ``Markerless gait analysis based on a single {RGB} camera,''
\newblock in {\em 15th {IEEE} International Conference on Wearable and
  Implantable Body Sensor Networks, {BSN} 2018, Las Vegas, NV, USA, March 4-7,
  2018}, 2018, pp. 42--45.

\bibitem{Grabcut}
Carsten Rother, Vladimir Kolmogorov, and Andrew Blake,
\newblock ``"grabcut": interactive foreground extraction using iterated graph
  cuts,''
\newblock {\em {ACM} Trans. Graph.}, vol. 23, no. 3, pp. 309--314, 2004.

\bibitem{MobileNet}
Andrew~G. Howard, Menglong Zhu, and etc,
\newblock ``Mobilenets: Efficient convolutional neural networks for mobile
  vision applications,''
\newblock {\em CoRR}, vol. abs/1704.04861, 2017.

\end{thebibliography}

\end{document}